\documentclass{article}

\usepackage{listings}
\usepackage{arxiv}
\usepackage{adjustbox}
\usepackage[utf8]{inputenc} 
\usepackage[T1]{fontenc}    
\usepackage{hyperref}       
\usepackage{url}            
\usepackage{booktabs}       
\usepackage{amsfonts}       
\usepackage{nicefrac}       
\usepackage{microtype}      
\usepackage{lipsum}		
\usepackage{graphicx}
\usepackage{natbib}
\usepackage{doi}
\usepackage{bm}
\usepackage{rotating}
\usepackage{tikz}
\usepackage{caption,color,float}

\usepackage[bb=px]{mathalfa} 

\usepackage[varvw]{newtxmath}       
\usepackage{amsmath}
\mathchardef\mhyphen="2D 
\usepackage{booktabs}
\usepackage{multirow}
\usepackage{comment}
\usepackage{diagbox}
\usepackage{soul}
\usepackage{fancyhdr}
\usepackage{wrapfig}
\usepackage{graphicx}

\usepackage{subcaption}

\usepackage{subcaption}

\title{Virtual Sensing-Enabled Digital Twin Framework for Real-Time Monitoring of Nuclear Systems Leveraging Deep Neural Operators}

\author{
    {Raisa Bentay ~Hossain} \\
    Nuclear, Plasma \& Radiological Engineering \\
    University of Illinois at Urbana-Champaign \\
    Urbana, IL 61801, USA \\
    \texttt{rh33@illinois.edu} \\
    \And
    {Farid ~Ahmed} \\
    Nuclear, Plasma \& Radiological Engineering \\
    University of Illinois at Urbana-Champaign \\
    Urbana, IL 61801, USA \\
    \And
    {Kazuma ~Kobayashi} \\
    Nuclear, Plasma \& Radiological Engineering \\
    University of Illinois at Urbana-Champaign \\
    Urbana, IL 61801, USA \\
    \texttt{kazumak2@illinois.edu} \\
    \And
    {Seid ~Koric} \\
    National Center for Supercomputing Applications \\
    University of Illinois at Urbana-Champaign \\
    Urbana, IL 61801, USA \\
    \texttt{koric@illinois.edu} \\
    \And
    {Diab ~Abueidda} \\
    National Center for Supercomputing Applications \\
    University of Illinois at Urbana-Champaign \\
    New York University Abu Dhabi \\
    United Arab Emirates \\
    \texttt{abueidd2@illinois.edu} \\
    \And
    {Syed Bahauddin ~Alam} \\
    Nuclear, Plasma \& Radiological Engineering \\
     National Center for Supercomputing Applications \\
    University of Illinois at Urbana-Champaign \\
    Urbana, IL 61801, USA \\
    \texttt{alams@illinois.edu} \\
}

\renewcommand{\headeright}{Preprint}
\renewcommand{\undertitle}{Preprint}
\renewcommand{\shorttitle}{\textit{Virtual Sensing-Enabled Digital Twin Framework for Nuclear Systems
}}

\hypersetup{
pdftitle={A template for the arxiv style},
pdfsubject={q-bio.NC, q-bio.QM},
pdfauthor={David S.~Hippocampus, Elias D.~Striatum},
pdfkeywords={First keyword, Second keyword, More},
}

\begin{document}
\maketitle

\begin{abstract}
{ \small

Effective real-time monitoring is a foundation of digital twin technology, crucial for detecting material degradation and maintaining the structural integrity of nuclear systems to ensure both safety and operational efficiency. Traditional physical sensor systems face limitations such as installation challenges, high costs, and difficulty measuring critical parameters in hard-to-reach or harsh environments, often resulting in incomplete data coverage. Machine learning-driven virtual sensors, integrated within a digital twin framework, offer a transformative solution by enhancing physical sensor capabilities to monitor critical degradation indicators like pressure, velocity, and turbulence. However, conventional machine learning models struggle with real-time monitoring due to the high-dimensional nature of reactor data and the need for frequent retraining. This paper introduces the use of Deep Operator Networks (DeepONet) as a core component of a digital twin framework to predict key thermal-hydraulic parameters in the hot leg of an AP-1000 Pressurized Water Reactor (PWR). DeepONet serves as a dynamic and scalable virtual sensor by accurately mapping the interplay between operational input parameters and spatially distributed system behaviors. In this study, DeepONet is trained with different operational conditions, which relaxes the requirement of continuous retraining, making it suitable for online and real-time prediction components for digital twin. Our results show that DeepONet achieves accurate predictions with low mean squared error and relative L2 error and can make predictions on unknown data \textit{1400 times faster than traditional CFD simulations}. This speed and accuracy enable DeepONet to synchronize with the physical system in real-time, functioning as a dynamic virtual sensor that tracks degradation-contributing conditions. By providing comprehensive system insights and predictive capabilities, DeepONet firmly establishes itself as a critical tool within the Digital Twin framework for nuclear systems.
}

\end{abstract}

%
{ 
\section{Introduction}
\label{sec:intro}

Proper monitoring and inspection of in-service components in nuclear reactors is essential for long-term safety and efficiency, as these components are continuously subjected to extreme temperatures, pressures, and radiation. Among these components, the primary circuit is particularly important, as it removes the immense heat generated in the reactor core, acting as the central component of the cooling system. This loop carries highly pressurized water at high velocities, creating significant mechanical stresses on the piping system \cite{Zou_20,Keller_02}. Turbulent coolant flow, especially in areas with bends and joints, leads to localized disturbances \cite{Tunstall_68,dutta_16, Shimizu_92} that can induce material degradation mechanisms such as erosion, fatigue, and stress corrosion cracking. Given the significant role of the primary coolant loop, real-time monitoring is essential for detecting early signs of degradation and preventing failures. Tracking key parameters such as pressure, velocity, and turbulence, enables the detection of deviations from normal operational conditions, such as flow reductions, vibrations, or pressure drops, which can serve as early indicators of potential material degradation. These insights provide early warnings, often before structural damage occurs, ensuring safe and efficient reactor operation while reducing risks associated with material degradation.


Non-destructive testing (NDT) methods, such as eddy-current, ultrasonic, radiographic, and visual inspections, are considered standard practices for degradation monitoring in nuclear reactors \cite{khan_23}. These methods effectively identify various forms of material degradation, including cracks, corrosion, and fatigue, without damaging the system itself. However, they are usually performed during scheduled outages, resulting in significant revenue losses. Additionally, nuclear piping systems often extend over several miles, making it challenging to inspect the entire system within a limited outage period \cite{Sandhu_24}. In-line inspections (ILI) are widely recognized as a standard approach for assessing internal pipeline conditions, including wall thinning, corrosion, and structural integrity \cite{integrity_2,integrity_1}. However, they are also conducted periodically and has similar issue as NDT methods. Thus, while traditional periodic inspection methods remain valuable, they present challenges in addressing material degradation during routine operations. Continuous monitoring of coolant flow in real time offers a promising alternative to scheduled inspections. By identifying anomalies in operational parameters as they occur, real time monitoring enables prompt scheduling of inspections to address specific concerns before they escalate into critical issues. Current pipeline monitoring systems use advanced simulation models to enable real-time monitoring of pressure conditions. These models offer valuable insights into operational parameters, particularly in areas where direct measurements are either impractical or impossible \cite{integrity_1}. However, simulation models take a long time to get results which make them unsuitable for continuous real-time monitoring tool. Real-time monitoring can address the shortcomings of periodic inspections if it can provide uninterrupted oversight of system parameters.

This uninterrupted view of entire reactor can be achieved through Digital Twin technology. A Digital Twin creates a virtual replica of the physical system, allowing for real-time monitoring, control, and prediction of system behavior \cite{kobayashi2024explainable,kobayashi2024deep,kobayashi2024improved}. By dynamically updating itself with real-time data, a Digital Twin provides a comprehensive and synchronized view of the reactor's operational state. This capability enables proactive maintenance, fault detection, and optimization of reactor operations, reducing the reliance on periodic inspection and minimizing operational risks. By continuously integrating operational data, Digital Twin frameworks enable predictive modeling of degradation processes, such as material fatigue and erosion-corrosion, ensuring timely interventions. Digital Twin framework rely heavily on sensors as they work as a connection between the physical reactor and its digital counterparts \cite{DT_sensor}. Sensors for coolant flow monitoring in nuclear reactors include ultrasonic, electromagnetic, and thermal dispersion flow meters \cite{Davydov_21,Tokarz_83}, as well as sensors for key parameters like temperature, pressure, and radiation \cite{Kuroze_96,Lister_15,Jensen_96}. These sensors are strategically placed at critical locations, such as the inlet and outlet of the reactor core, steam generators, and other components, to ensure continuous monitoring of system conditions \cite{NRC2011}, as well as to preserve sensor health and minimize degradation \cite{hossain2024sensor, niharika_24}. However, for a successful Digital Twin framework, comprehensive coverage of the entire reactor system is needed, which cannot be achieved solely through physical sensors.  While placing sensors at the inlet and outlet of piping segments is feasible, installing them within pipe sections is impractical as it will cause flow disturbances and also has logistical challenges. Though DFO can address coverage limitations, it is very expensive and cannot be implemented in existing reactors.

Virtual sensors can address these limitations of physical sensors. They are software-based models that estimate physical quantities using data and simulations rather than direct physical measurements \cite{zhao2024virtual,zhao2024graph,niresi2024physics}. They mimic the behavior of physical sensors, providing readings without requiring the installation of actual hardware \cite{sun_2017, Masti_21, Ofner_23}. Integrated within a Digital Twin, virtual sensors utilize data from sporadically placed physical sensors to predict values in unseen or unmonitored areas, providing critical insights into system conditions in real time and offering a complete view of the reactor's operational state. Unlike physical sensors, virtual sensor-based Digital Twin frameworks are not constrained by installation or environmental limitations, making them ideal for monitoring hard-to-reach or harsh reactor environments \cite{Ahmed_2013}. Furthermore, Digital Twin frameworks utilize existing sensor networks to operate virtual sensors, reducing the need for modifications to established reactor designs while still expanding monitoring capabilities. They also improve data reliability by reducing signal interference and are easily adjustable to accommodate changes in system requirements \cite{liu_09, VS_webpage}. This adaptability ensures the Digital Twin remains accurate and reliable, even when physical sensors degrade or fail.

While virtual sensing is still a relatively new concept in nuclear applications, its potential to enhance system reliability and efficiency has been demonstrated over past years. For instance, Sevilla \textit{et al.} \cite{sevilla_98} showed how neural networks could estimate variables in pressurized water reactors by optimizing input selection and network architecture. Ahmed \textit{et al.} \cite{Ahmed_2013} developed virtual sensor networks for accident monitoring in nuclear plants, while Tipireddy \textit{et al.} \cite{tipireddy_17} introduced Gaussian process-based virtual sensors to replace faulty physical sensors, reducing unscheduled downtime. These examples underscore the critical role virtual sensing plays in Digital Twin frameworks, enabling real-time monitoring, fault tolerance, and system optimization in nuclear reactors.

Emerging neural network technologies, such as the Deep Operator Network (DeepONet) \cite{lulu_21}, Fourier Neural Operators (FNO) \cite{FNO_20}, and Physics-Informed Neural Networks (PINN) \cite{PINN_19} provide powerful tools for developing virtual sensors. These models are adept at solving partial differential equations (PDEs) by approximating nonlinear operators and learning mappings between functional spaces. This makes them highly suitable for modeling and predicting the behavior of complex, interdependent systems, such as nuclear reactors. Among these algorithms, the Deep Operator Network (DeepONet) has demonstrated greater efficiency compared to others. Lu \textit{et al.} in their recent paper \cite{Lulu_22} show that, DeepONet is more flexible than FNO in terms of problem settings and datasets. With PINN, as well, each new parameter requires separate simulations or retraining. Moreover, PINN struggles to approximate PDEs with strong non-linearity commonly found in practical fluid flow problems. In contrast, DeepONet alleviates the need for retraining by learning operators that map entire functions, rather than specific input-output pairs, across different conditions. Once trained, DeepONet can generalize to a wide range of new inputs without requiring retraining for each new scenario, as it understands the underlying functional relationships between inputs and outputs. This is advantageous for real-time applications, adapting to varying conditions without frequent retraining. DeepONet approximates linear and nonlinear PDE solution operators using parametric functions as inputs, mapping them to output spaces, thereby eliminating retraining requirements \cite{koric_23}.

A well-trained neural operator offers the computational efficiency needed for real-time or near real-time predictions, which is crucial for control system optimization \cite{kobayashi2024improved, diab_24}. These advantages make DeepONet our preferred algorithm for establishing Digital Twin. Unlike traditional finite element/volume (FEM/FVM) simulations, which are computationally intensive and time-consuming, DeepONet generates predictions orders of magnitude faster, making it ideal for real-time applications. By training on high-fidelity simulations and experimental data, DeepONet accurately predicts coolant flow behavior under various operating conditions, which is central to developing a virtual sensor framework for nuclear reactors. The key advantage of using DeepONet for our problem is its trunk network, which processes spatial coordinates to evaluate the output function \cite{lulu_21, diab_24}. This network predicts parameters at various pipe locations based on input data from a single location, providing comprehensive pipe condition inference. Combined with near real-time capability, DeepONet provides highly accurate estimates of flow velocity, turbulence, pressure, and temperature at key primary circuit locations without physical sensors.

\begin{figure}[htbp]
    \centering
    \includegraphics[width=0.35\textwidth]{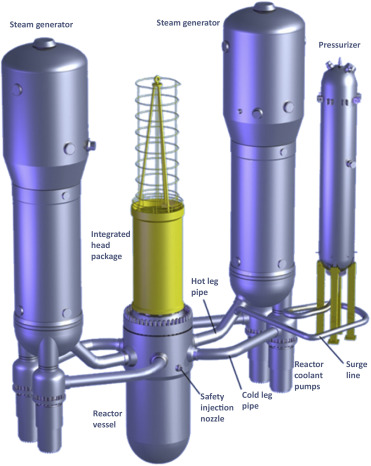}
\caption{Schematic diagram of AP-1000 reactor coolant system \cite{Namgung_16}.}
\label{fig:ap_100}
\end{figure}

In this work, we study the hot leg as the representation of the primary coolant system to demonstrate DeepONet's effectiveness. The hot leg, consisting of horizontal and vertical conduits with elbows, transports high-pressure, high-temperature water from the reactor pressure vessel (RPV) to the steam generator (SG). Leakage in the hot leg can cause serious issues, including a loss of coolant accident (LOCA), impairing reactor's cooling capacity and posing radiological risks. Monitoring coolant conditions in the hot leg is crucial to ensure proper heat transfer and reactor safety. However, real-time monitoring of key parameters like pressure, velocity, and turbulence kinetic energy inside the hot leg is challenging due to critical operating conditions. We focus on developing a real-time monitoring model of thermal-hydraulic conditions for the hot leg of the AP1000, as shown in Figure \ref{fig:ap_100}. We use the coolant inlet velocity, already monitored through existing sensors, assuming normal PWR operational conditions. Based on this information, we develop a model to predict thermal-hydraulic parameters (pressure, velocity, and turbulence kinetic energy) in the hot leg's central plane. Monitoring these parameters is important for assessing heat transfer performance and ensuring structural integrity. For instance, high or fluctuating pressure stresses piping, causing wear, fatigue, and cracking, while sudden drops in pressure may signal blockages or leaks, accelerating degradation.
Monitoring velocity is equally important, as high coolant flow velocity, particularly in areas where the flow direction changes, can cause flow-induced vibrations. These vibrations lead to mechanical fatigue and increase the risk of erosion and corrosion, gradually thinning the pipe walls. Similarly, turbulence, especially near bends and joints, amplifies stress on the piping structure, accelerating fatigue and the erosion-corrosion process. It also introduces hydraulic shocks and uneven temperature distributions, further weakening material integrity over time. Monitoring these parameters in real time enables operators to correlate abnormal flow patterns with potential degradation, allowing for timely maintenance decisions. Given their role in degradation, our goal is to develop a real-time thermal-hydraulic monitoring model for the hot leg of a PWR.

The key contribution of our work is demonstrating the feasibility and benefits of using DeepONet within a Digital Twin framework for real-time monitoring of coolant flow in the primary circuit of a nuclear reactor. By integrating virtual sensors, our study highlights the potential of Digital Twins to significantly improve the monitoring and inspection of reactor components. Through comprehensive simulations and experimental validation, we show how this approach delivers accurate and timely data, enabling the detection of operational conditions indicative of degradation and facilitating proactive maintenance and optimizing reactor operations.

\section{Methods}
\subsection{Data Generation}
The hot leg in an AP1000 reactor plays a critical role in its operation, acting as the primary channel for transferring heat generated in the reactor core to the steam generators. Hence, the core could overheat without proper flow through the hot leg, leading to a potential meltdown. AP1000 reactor cooling system includes hot and cold leg pipes connecting the reactor vessel, steam generators, and reactor coolant pumps. Each loop has three pipes, including a 787.4 mm inner diameter pipe between the reactor vessel outlet and steam generator inlet with a length of 2.3 m \cite{Namgung_16}. While system-level codes provide a useful tool for macroscopic thermal-hydraulic analysis of nuclear reactors, their limitations in resolving detailed flow features such as flow re-circulation and turbulence within the hot leg necessitate the use of more advanced methods like Computational Fluid Dynamics (CFD). CFD, with its ability to resolve complex geometries and flow physics, can provide a more accurate and detailed understanding of the fluid dynamics within the hot leg. However, due to the complexity of the full-scale geometry, a scaled-down model was used for CFD analysis. Geometric scaling was employed, ensuring a constant flow rate per unit volume between the actual case and the model. In this study, geometric scaling was implemented according to the following conditions:

\begin{equation*}
    \frac{Q_m}{V_m} =  \frac{Q_a}{V_a}
\end{equation*}

 Where , $Q_m$ and $Q_a$ are the volumetric flow rates of model and actual case and $V_m$ $V_a$ are the volume which results in volume of the model and actual case. Hence we can write,   
 
\begin{equation*}
    \frac{v_m}{v_a} =  \frac{l_m}{l_a}
\end{equation*}

where $v_m$ and the $v_a$ are the velocities of the model and the actual case. $l_m$ and $l_a$ represents the model and actual flow length the hot leg.The diameter of the hotleg pipe ($d_a$) was scaled down by the a factor of $\lambda$ equivalent to 31.5 which gives the value of $d_m$ to be 25 mm. The flow length of the model was kept 150 mm where $l_a = d_m \times l_m$. Hence. the relationship between the Reynolds number (Re) for the model and actual scenario becomes:

\begin{equation*}
    Re_m = Re_a \times \frac{\lambda}{d_m}
\end{equation*}

As shown in Figure \ref{geom_bc}, the elbow joint angle ($\theta_c$) was kept $120^{\circ}$, same as the actual scenario of hot leg.

The walls of the flow domain were treated as adiabatic and subject to no-slip conditions. This implies that these surfaces neither gained nor lost heat, and fluid velocity at the wall interface was zero.  In terms of thermal conditions, the inlet fluid temperature was maintained at 594.3 K.  The outlet was set to a gauge pressure of zero, ensuring it remained at ambient atmospheric pressure. The inlet velocity ($v_{in} $) was changed according to the magnitude of Re , ensuring turbulent flow. Figure \ref{geom_bc} shows the essential boundary conditions and the plane of interest where the hydrodynamics parameters evaluated through CFD simulations to train ML model.

In the numerical investigations, the coolant was modeled as a Newtonian fluid with constant viscosity and density, implying a linear relationship between stress rate and strain rate.  Simulations were performed under steady-state conditions, encompassing both forced turbulent flow regimes.3D Navier-Stokes equations solved using the Finite Volume Method (FVM). The governing equations are as follows \cite{ahmed2024enhancing}: 

Continuity equation:

$$
\frac{\partial \rho_{\mathrm{}}}{\partial t}+\nabla \cdot\left(\rho_{\mathrm{}} \vec{v}\right)=0
$$

Momentum equation:

$$
\frac{\partial\left(\rho_{\mathrm{}} \vec{v}\right)}{\partial t}+\nabla \cdot \rho_{\mathrm{}} \vec{v} \vec{v}=-\nabla P+\nabla \overline{\bar{\tau}}+\rho_{\mathrm{}} g
$$

Energy equation:

$$
\frac{\partial(\rho H)}{\partial t}+\nabla \cdot(\rho H \vec{v})=-\nabla \cdot \vec{q}^{\prime \prime}+q^{\prime \prime \prime}-P \nabla \cdot \vec{v}
$$

In the present study, forced turbulent flows were assessed employing the RNG $k-\varepsilon$ model, the equations for which are presented below \cite{ahmed2022assessment,ahmed2021thermohydraulic}:

\begin{equation*}
    \frac{\partial}{\partial t}(\rho k)+\frac{\partial y}{\partial z_{\mathrm{i}}}\left(\rho \varepsilon u_{\mathrm{i}}\right)=\frac{\partial}{\partial z_{\mathrm{j}}}\left(\alpha_{\mathrm{k}} \mu_{\mathrm{eff}} \frac{\partial k}{\partial z_{\mathrm{j}}}\right)+G_{\mathrm{k}}+G_{\mathrm{b}}-\rho \varepsilon-Y_{\mathrm{M}}+S_{\mathrm{k}}\\
\end{equation*}

\begin{equation*}
    \frac{\partial}{\partial t}(\rho \varepsilon)+\frac{\partial y}{\partial x_{\mathrm{i}}}\left(\rho \varepsilon u_{\mathrm{i}}\right)=\frac{\partial}{\partial z_{\mathrm{j}}}\left(\alpha_{\mathrm{k}} \mu_{\mathrm{eff}} \frac{\partial k}{\partial z_{\mathrm{j}}}\right) \\
   \quad+C_{1 \varepsilon} \frac{\varepsilon}{k}\left(G_{\mathrm{k}}+C_{3 \mathrm{k}} G_{\mathrm{b}}\right)-C_{2 \varepsilon} \rho \frac{\varepsilon^2}{k}-R_{\varepsilon}+S_{\varepsilon}
\end{equation*}

\begin{figure}[htbp]
    \centering
\includegraphics[width=0.8\textwidth]{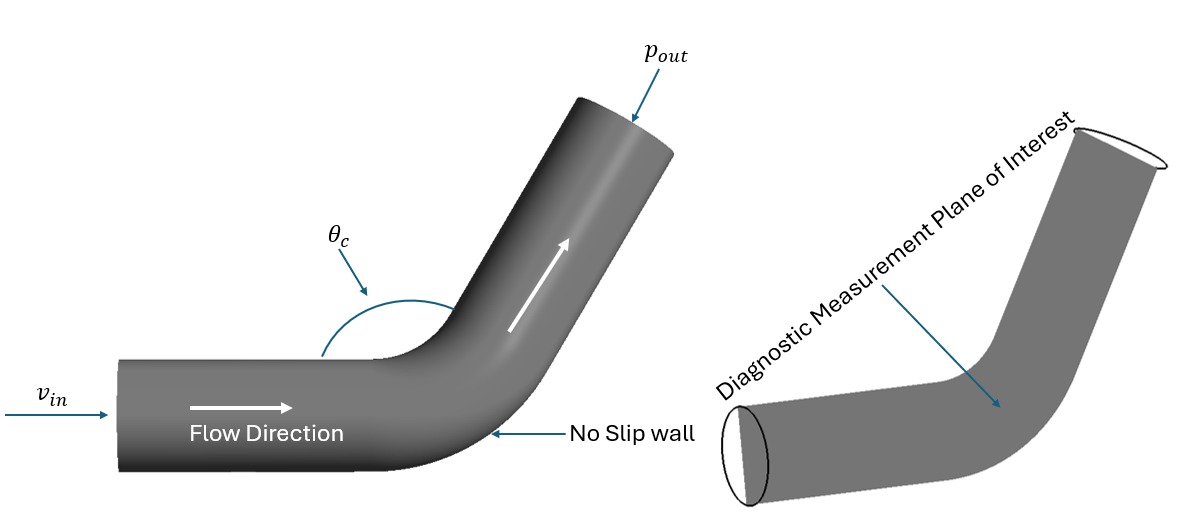}
\caption{Computational domain and boundary conditions}
\label{geom_bc}
\end{figure}

In the FVM, the establishment of a computational grid over the geometry is essential for attaining solution convergence. To facilitate this, the present study employed hexahedral and tetrahedral mesh spanning the entire computational domain, with particular attention dedicated to enhanced wall treatment for improved accuracy.   Mesh with refined wall treatment, including ten boundary layers, was employed to ensure solution convergence.  Mesh elements were finely sized at 3.75 mm near critical regions.  Mesh quality metrics were strictly controlled, with skewness and orthogonal quality maintained at 0.11 and 0.96, respectively, across all channel configurations. Figure \ref{fig:mesh} represents the grid generation over the fluid domain.

\begin{figure}[htbp]
    \centering
    \includegraphics[width=0.8\textwidth]{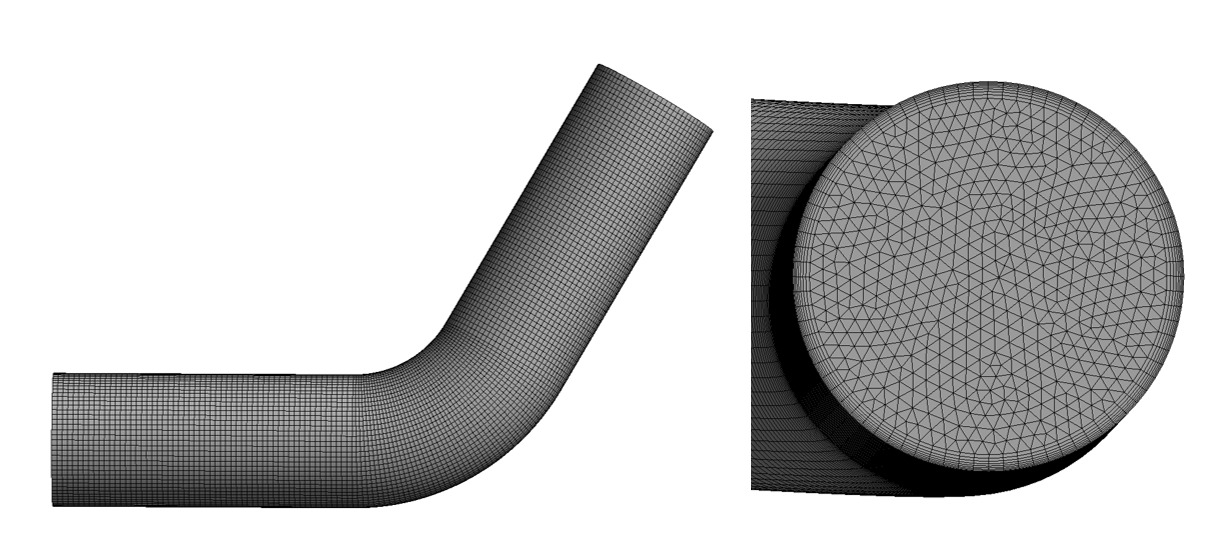}
\caption{Grid generation over the domain}
\label{fig:mesh}
\end{figure}

This study utilized an implicit time-marching scheme to solve the governing equations. The pressure-velocity coupling was achieved through the SIMPLEC algorithm. For the discretization of mass, momentum, and energy conservation equations, a second-order upwind scheme was implemented, balancing accuracy and computational efficiency. The turbulence kinetic energy and dissipation rate equations were discretized using a first-order upwind scheme, chosen for its robustness and stability in turbulent flow simulations. The turbulence intensity ($ I $) at the inlet was specified based on the following empirical relationship \cite{homicz2004computational}:

\begin{equation*}
    I \equiv \frac{\sqrt{\frac{2}{3} k}}{U_{\mathrm{avg}}} \cong \frac{0.16}{R e^{1 / 8}}
\end{equation*}

\subsection{DeepONet Architecture} 

In this study, we have used an unstacked DeepONet described by Lu Lu \textit{et al.} \cite{lulu_21}, consists of a single branch network and a trunk network. Figure \ref{fig:model_arc} shows the model architecture used in this work. The DeepONet architecture lays the foundation for capturing the interplay between dynamic operational input parameters and spatially distributed system behaviors (spatial coordinates), a core characteristic of digital twin frameworks.  The \textit{branch} network processes the input function, which, in our case, represents the average initial velocity. This input, belonging to an infinite-dimensional functional space, is characterized by $n$ control points. Here, $n=$1, as we are considering the average value. The \textit{trunk} network, on the other hand, handles the spatial information. It takes as input a collection of $N$ points within the domain, each defined by its $(x, y, z)$ coordinates. These points correspond to 11,340 virtual sensors ($N$=11,340) distributed throughout the pipe section. By encoding spatial dependencies and integrating dynamic operational data, the trunk network ensures that the digital twin remains synchronized with the physical system. 

The core capability of DeepONet, which justifies its role as a digital twin, lies in the fusion of real-time information from the branch network and spatial information from trunk networks. This integration is achieved through element-wise multiplication \cite{Jin_22, he_2023} of the outputs from Branch and Trunk Networks. The resulting output, $G(u)(y)$, is a function of the spatial coordinates $y$ conditioned on the input function $u$. This formulation enables the network to learn complex mappings between the input and output spaces, empowering real-time predictions of critical thermal-hydraulic parameters such as pressure, velocity, and turbulence kinetic energy. This aligns with the NRC's definition \cite{yadav2023state} of a digital twin, which emphasizes real-time synchronization, predictive capabilities, and comprehensive system representation.

\begin{figure}[htbp]
    \centering
    \includegraphics[width=0.9\textwidth]{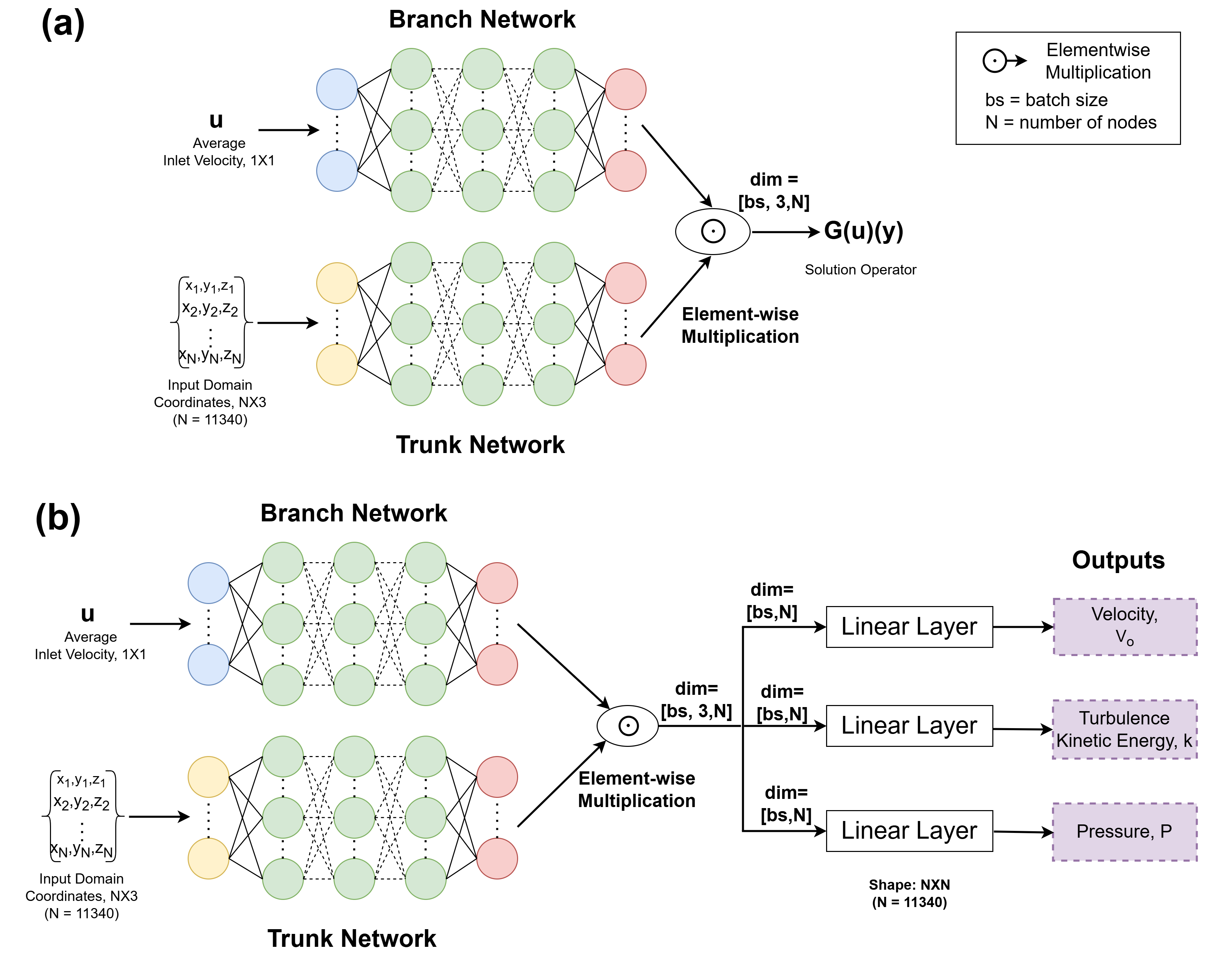}
\caption{Schematic of the FNN-based DeepONet architecture used in this study. The model consists of a single branch and trunk network. The branch network takes the average inlet velocity (\(u\)) as input, and the trunk network takes the spatial domain coordinates (\(x, y, z\)). The output quantities are distributions of coolant pressure (\(P\)), velocity (\(V_o\)), and turbulence kinetic energy (\(k\)). The schematic (a) shows the original DeepONet architecture. The Branch network has 11 hidden layers and the trunk network has 10 hidden layeres with 4096 neurons each. (b) The schemetic  illustrates the modified architecture with additional linear layers for each parameter. The branch network has layer sizes of \([n, 512, 512, 512, N]\) where \(n=1\), and the trunk network has layer sizes of \([3, 512, 512, 256, 3]\), both utilizing ReLU activation. At the end, there are three linear layers, each with sizes \([N, N]\), where \(N=11,340\), without any activation.
}
\label{fig:model_arc}
\end{figure}

To enhance the model's predictive capabilities, we added additional linear layers into the original DeepONet architecture. The concept of utilizing linear layers was inspired by the work of Kazuma \textit{et al.} \cite{Kazuma_24_M}. These linear layers consist of three independent layers, each dedicated to refining one of the predicted parameters - velocity, turbulence kinetic energy, or pressure. Each linear layer takes an input vector corresponding to the spatial domain (in our case 11,340 spatial points) and outputs a vector of the same size, ensuring that the predictions retain the spatial resolution required for monitoring. The weights of these layers are initialized using Xavier initialization, which ensures stable gradients and efficient training, while biases are initialized to zero to prevent unnecessary offsets in the early stages. This modular design allows each linear layer to specialize in its respective parameter, enabling parameter-specific transformations that improve prediction accuracy. These layers enable scaling and shifting of the output, improving alignment with the target values. Figure \ref{fig:model_arc} displays the original model architecture introduced by Lu Lu \textit{et al.} \cite{lulu_21} (a), alongside the modified architecture used in this study (b). After the element-wise multiplication of the outputs from the Branch and Trunk Networks, the resulting tensor is split into three separate components, with each component corresponding to a specific parameter (velocity, turbulence, or pressure). Each component is then passed through its respective linear layer, which applies a final transformation. For a batch of data, the input tensor to the linear layers has dimensions \textit{(batch-size, no-of-parameters, number-of-spatial-points)}. After processing, each linear layer outputs a tensor of size \textit{(batch-size, number-of-spatial-points)}, in our case $(512,11340)$,  preserving spatial resolution by mapping input vectors directly to corresponding output vectors. In section \ref{comp}, we present a comparison between the models with and without the additional linear layers to further justify this modifications. The DeepONet model was trained to predict three primary output functions: turbulence, pressure, and velocity distributions in the central plane of the hot leg. These output functions reside in distinct functional spaces, denoted as \( S_1 \), \( S_2 \), and \( S_3 \), where \( k \in S_1 \), \( p \in S_2 \), and \( v_o \in S_3 \) represent turbulence, pressure, and velocity values in the central plane. And the mapping solution operators of the DeepONet can be defined as \( G_1 \), \( G_2 \), and \( G_3 \) for three different parameters.

For both branch and trunk networks, we employed Feedforward Neural Networks (FNNs) due to their simplicity and effectiveness in approximating nonlinear functions. The architecture of these FNNs included three hidden layers, with the number of neurons in each layer determined through hyperparameter tuning. This hyperparameter optimization process was crucial in achieving optimal performance. The activation function of choice for the branch and trunk networks was the Rectified Linear Unit (ReLU). The Adam optimizer, known for its efficiency and stability, was employed to update the network parameters iteratively. The training process involved minimizing the scaled Mean Squared Error (MSE) loss function:

\begin{equation}
\label{eq:mse}
    \text{MSE Loss} = \frac{1}{N} \sum_{i=1}^{N} (y_i - \hat{y}_i)^2
\end{equation}

where \( N \) presents the number of data points (no of nodes), \( y_i \) is the predicted value, and \( \hat{y}_i \) is the true value.

\section{Results and Discussion}
\label{result_discussion}

\subsection{Data Processing} 
The data for this study was generated using ANSYS Fluent, focusing on fluid dynamics within the hot leg elbow joint of an AP-1000 LWR nuclear reactor. The average inlet velocity range for the simulations was set between 0.63 and 0.83. The computational mesh consisted of a total of 11,340 nodes, distributed along the central plane of the hot leg to capture the coolant flow characteristics throughout the pipe section. For this study, we performed a total of 5,000 simulations. This dataset was subsequently divided into a training and testing set in a 80\%-20\% ratio, resulting in 4,000 scenarios for training and 1000 scenarios for testing. Importantly, the 1000 test data points were never touched during the training or validation process. They remained completely unseen by the model and were used exclusively at the final stage to evaluate the model's performance on unseen data. This setup was designed to monitor coolant behavior in near-real-time, capturing detailed velocity, turbulence, and pressure variations to detect any anomalies effectively.

To ensure the DeepONet model properly learns the values of different parameters, we addressed the issue of varying ranges among these parameters. For instance, the pressure values range between approximately -231.25 and 132.7, while the turbulence kinetic energy values range between approximately 0.000875 and 0.019015. Such discrepancies in value ranges can hinder the model's learning process. To mitigate this difference, we applied min-max scaling to normalize all the values between 0 and 1. 

\begin{equation}
X_{\text{scaled}} = \frac{X - X_{\text{min}}}{X_{\text{max}} - X_{\text{min}}}
\end{equation} 

where $X$ is the original value, $X_{\text{min}}$ is the minimum value in the dataset, $X_{\text{max}}$ is the maximum value in the dataset, and $X_{\text{scaled}}$ is the normalized value in the range [0, 1].

\begin{figure}[h!]
    \centering
    \includegraphics[width=\textwidth]{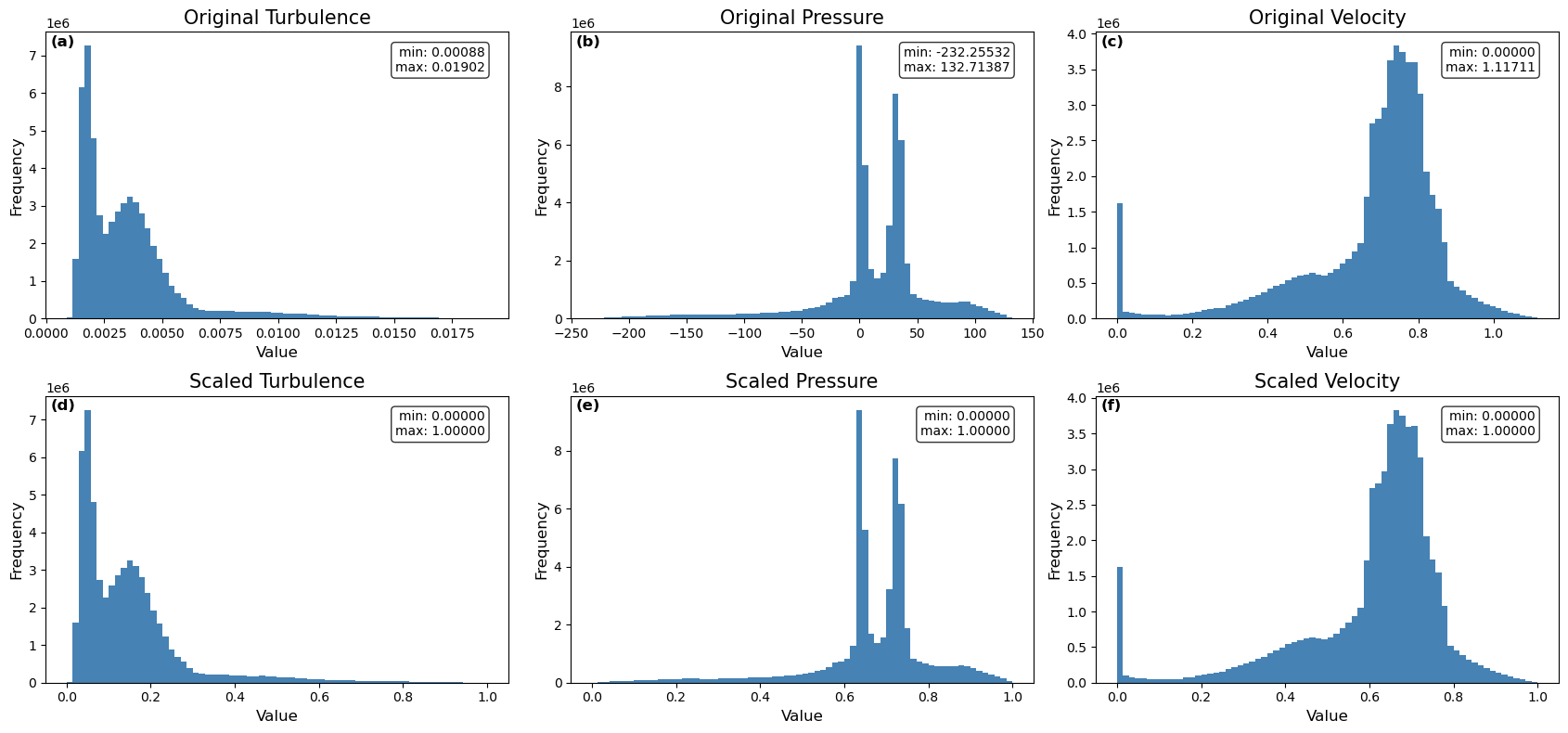}
\caption{Histograms of original and scaled parameter values. The top row displays the original values for turbulence, pressure, and velocity, showing distinct distributions. The bottom row presents the values after min-max scaling, normalizing all parameters to the [0,1] range.}
\label{fig:scaling}
\end{figure}

Figure \ref{fig:scaling} provides a visual representation of this scaling process. The histograms in the top row display the original values of kinetic turbulence energy, pressure, and velocity. The bottom row shows the scaled values of these parameters after applying min-max scaling. From these histograms, it is evident that the original values of turbulence (a), pressure (b), and velocity (c) are distributed across different ranges. The original turbulence values are densely packed near the lower end of the range, with a significant peak around 0.0025. The pressure values exhibit a broader distribution with notable peaks around -50 and 50. The velocity values are concentrated between 0.6 and 0.8, with a pronounced peak near 0.75. After scaling, the turbulence values (d) are now spread across the [0,1] range, maintaining the original distribution shape but normalized. The scaled pressure (e) and velocity (f) values are also within the range of [0,1], preserving the distribution characteristics seen in the original data. By scaling these parameters, we ensure that the DeepONet model can learn effectively from the data, with each parameter contributing proportionately to the training process. This step helps improving the model's accuracy and reliability in predicting coolant behavior.\\

\subsection{Training Process} 

The training process for this study is designed to ensure that the model fits the training data well while also generalizing effectively to unseen data. Initially, the training dataset is split into two parts: 80\% for training and 20\% for validation. The final model, including the linear layer, was trained for 100 epochs to ensure convergence with this specific dataset, activation function, and optimizer. The number of hidden layers for the final model was fixed at three, but the number of neurons in each layer was determined through hyperparameter tuning.

Hyperparameter tuning for the final model with linear layers was achieved through Bayesian optimization using the Optuna framework, which efficiently searches for optimal parameters. The loss function employed was Mean Squared Error (MSE), and the model was optimized using the Adam optimizer with L2 regularization to prevent overfitting. Training was conducted over 100 epochs, with early stopping implemented to halt training if performance did not improve within 5 consecutive epochs, ensuring efficient use of computational resources. The model underwent 5-fold cross-validation to assess its generalization capabilities across different subsets of the dataset. This process ensures that each data point is used for both training and validation, providing a comprehensive evaluation of the model's performance. The average performance across all five folds gives a reliable estimate of how well the model will generalize to new data. The total training duration was 19 hours and 38 minutes. The computations for the training and evaluation tasks were performed on a computational node with a single NVIDIA A100 GPU within the Delta cluster hosted by the National Center for Supercomputing Applications (NCSA). While hyperparameter tuning determined most parameters, the number of linear layers was decided through manual inspection based on observed improvements in performance. Table \ref{table:hyperparameters} summarizes the ranges and best values found for each parameter.

\begin{table}[htbp!]
\caption{Summary of Hyperparameter Tuning for DeepONet with linear layers.}
\centering
\label{table:hyperparameters}
\renewcommand{\arraystretch}{1}
\resizebox{0.65\textwidth}{!}{%
\begin{tabular}{@{}l|l|l@{}}
\hline 
\textbf{Parameter Tuned} & \textbf{Range} & \textbf{Best Value Found} \\ \hline 
No of Neurons for each hidden layer & 128, 256, 512       & 512      \\ \hline 
Dropout Rate      & 0.0 to 0.3                               & 0.0013836 \\ \hline 
Learning Rate     & \(10^{-5}\) to \(10^{-3}\)   & 0.00098 \\ \hline 
Batch Size        & 16, 32, 64                               & 16               \\ \hline 
Weight Decay Rate & \(10^{-8}\) to \(10^{-6}\)   & 1.53625e-08 \\ \hline 
Number of Trials  & 50                                       & NAN              \\ 
\hline 
\end{tabular}%
}
\end{table}

After confirming that the model generalizes well through cross-validation, the next step is to train the model on the entire dataset for an extended period of 1000 epochs. For this final training phase, the hidden layer configuration was [512, 512, 512], the dropout rate was set to 0.0, the weight decay rate was \(10^{-8}\), and the learning rate was 0.001.
Training on the whole dataset allows the model to learn from all available data, maximizing its potential to capture complex patterns. The extended training period ensures that the model has sufficient time to converge to an optimal solution.

After training the model on the entire dataset, the best model is then used to make predictions on the test dataset. The test dataset consists of data points that were not used during training or validation, providing an unbiased evaluation of the model's performance. These 1000 test data points remained completely unseen by the model and were used exclusively at this stage to evaluate the model's performance on unseen data. To evaluate the model on test dataset following quantities were calculated:

\begin{equation}
\text{Mean Square Error (MSE)} = \frac{1}{N} \sum_{i=1}^{N} (y_i - \hat{y}_i)^2
\end{equation}

\begin{equation}
\text{Mean Absolute Error (MAE)} = \frac{1}{N} \sum_{i=1}^{N} |y_i - \hat{y}_i|
\end{equation}

\begin{equation}
\text{Relative L2 Error} = \frac{\| y_i - \hat{y}_i \|_2}{\| y_i \|_2}
\end{equation}

where \(y_i\) represents the true values generated through finite volume method, \(\hat{y}_i\) represents the predicted values from neural network, and \(N\) is the total number of data points.

The values reported in the tables are calculated as the averages over the 1000 test scenarios in the test set, ensuring an unbiased evaluation of model performance. The average values for Mean Square Error (MSE), Mean Absolute Error (MAE), and Relative L2 Error are computed as follows:

\begin{equation}
\text{Average Metric}  = \frac{1}{M} \sum_{j=1}^{N} \text{Metric}_j
\end{equation}

Here, \(\text{Metric}_j\) is the value of the metric (MSE, Relative L2 Error, MAE) for the \(j\)-th test scenario. And \(M\) is total number of test scenarios (1000 in this case). The formula aggregates the metric for all test scenarios and divides it by the total number of scenarios to compute the average.

\subsection{Vanilla DeepONet vs DeepONet with Linear Layers} \label{comp}

To support our modification of the original model with added linear layers for each parameter separately, we present a comparison between the DeepONet model with and without these linear layers. This comparison is conducted purely to evaluate the impact of adding linear layers, and the Vanilla DeepONet model is used solely for this purpose.

To match the total number of learnable parameters for both of the model, the number of hidden layers for Vanilla DeepONet model was taken to be 11 for Branch network and 10 for Trunk network. Each hidden layer has 4096 neurons. Which makes the total number of learnable parameters to be 365431887. On the other hand, our considered model with linear layer has 3 hidden layers with 512 neurons each. With the extra three linear layers at the end, the total number of learnable parameters for this model is 361499727. With that these two models become comparable with each other. Table \ref{table:deeponet_comparison} shows the average value, standard deviation, and maximum value of the mean square error (MSE) and relative L2 error on the test dataset for each model, evaluated separately for the three parameters: pressure, velocity and turbulence kinetic energy. It is important to note that no hyperparameter tuning was performed for the Vanilla DeepONet model. The parameters for this model were taken to match those of the DeepONet model with linear layers, as described in Section 3.2. Since the Vanilla DeepONet model is not used for further evaluation or visualization, hyperparameter tuning was not necessary.

\begin{table}[h!]
\centering
\renewcommand{\arraystretch}{1.5}
\resizebox{\textwidth}{!}{
\begin{tabular}{c|c|c|c|c|c}
\hline
{} &{}                     & \multicolumn{2}{c|}{Mean Square Error (MSE)}             & \multicolumn{2}{c}{Relative L2 Error}            \\ \cline{3-6} 
{\multirow{-2}{*}{Parameters}} &{\multirow{-2}{*}{Model}} & Average (Std)           & Maximum Value & Average (Std)         & Maximum Value \\ \hline
\multirow{2}{*}{Pressure (P)}       & Original DeepONet  & 5.00$\times10^{-4}$(2.30$\times10^{-4}$)       & 1.02$\times10^{-3}$       & 3.17$\times10^{-2}$(\textbf{7.37$\times10^{-3}$})     & 4.68$\times10^{-2}$       \\ \cline{2-6} 
                            & With Linear Layers & \textbf{2.60$\times10^{-4}$}(2.30$\times10^{-4}$)       & \textbf{8.80$\times10^{-4}$}       & \textbf{2.01$\times10^{-2}$}(1.18$\times10^{-2}$)     & \textbf{4.28$\times10^{-2}$}       \\ \hline 
\multirow{2}{*}{Velocity ($V_o$)}     & Original DeepONet  & 4.02$\times10^{-3}$(1.52$\times10^{-3}$)        & 8.31$\times10^{-3}$       & 1.00$\times10^{-1}$(\textbf{2.30$\times10^{-2}$})     & 1.61$\times10^{-1}$        \\ \cline{2-6} 
                            & With Linear Layers & \textbf{1.40$\times10^{-3}$}(1.27$\times10^{-3}$)       & \textbf{4.59$\times10^{-3}$}       & \textbf{5.14$\times10^{-2}$}(3.01$\times10^{-2}$)     & \textbf{1.09$\times10^{-1}$}       \\ \hline
\multirow{2}{*}{Turbulence Kinetic Energy (k)} & Original DeepONet  & 1.46$\times10^{-3}$(5.20$\times10^{-4}$)       & 2.96$\times10^{-3}$      & 1.95$\times10^{-1}$(\textbf{3.30$\times10^{-2}$})     & 2.84$\times10^{-1}$       \\ \cline{2-6} 
                            & With Linear Layers & \textbf{5.30$\times10^{-4}$}(4.80$\times10^{-4}$)       & \textbf{1.81$\times10^{-3}$}       & \textbf{1.03$\times10^{-1}$}(6.05$\times10^{-2}$)     & \textbf{2.33$\times10^{-1}$}       \\ \hline
\end{tabular}
}
\caption{Comparison of Mean Square Error (MSE) and Relative L2 Error for Original DeepONet and DeepONet with additional Linear Layers.}
\label{table:deeponet_comparison}
\end{table}

Analyzing the data in Table \ref{table:deeponet_comparison}, it is evident that the model with linear layers consistently outperforms the original DeepONet model across all parameters. For pressure (P), the MSE average decreases from 0.00048 to 0.00026, and the relative L2 error average drops from 0.03106 to 0.0204, indicating a significant improvement in predictive accuracy. Similarly, for velocity ($V_o$), the model with linear layers shows a reduction in the MSE average from 0.00281 to 0.00141, and a decrease in the relative L2 error average from 0.07934 to 0.05184. Turbulence kinetic energy (k) also benefits from the linear layers, with the MSE average reducing from 0.00105 to 0.00054, and the relative L2 error average declining from 0.15924 to 0.10573. These enhancements demonstrate the efficacy of the added linear layers in refining the model's alignment with target values, thus validating our modifications.\\

\subsection{Impact of Data Splits and Node Counts on Model Accuracy} 

To evaluate the impact of different train-test splits on the model’s performance, the dataset was divided into the following ratios: 70-30\%, 80-20\%, and 90-10\%. The model was trained on these varying fractions of the training set, and the average Mean Squared Error (MSE) and Relative L2 error were calculated for each test split. It was hypothesized that the model's performance would degrade with a smaller training dataset, as less data typically leads to poorer model performance. However, the results were surprisingly consistent across different splits, as shown in Table \ref{table:split_results}.

\begin{table}[h]
\centering
\renewcommand{\arraystretch}{1}
\resizebox{\textwidth}{!}{%
\begin{tabular}{c|c|c|c|c|c|c}
\hline
\multirow{2}{*}{\textbf{Train-Test Split}} & \multicolumn{3}{c|}{\textbf{Average MSE}} & \multicolumn{3}{c}{\textbf{Average Relative L2 Error}} \\
\cline{2-7}
 & \textbf{Pressure} & \textbf{Velocity} & \textbf{Turbulence} & \textbf{Pressure} & \textbf{Velocity} & \textbf{Turbulence} \\
\hline
90\%--10\% & 2.59$\times10^{-4}$ & 1.41$\times10^{-3}$ & 5.39$\times10^{-4}$ & 2.04$\times10^{-2}$ & 5.18$\times10^{-2}$ & 1.06$\times10^{-1}$ \\
\hline
80\%--20\% & 2.55$\times10^{-4}$ & 1.40$\times10^{-3}$ & 5.28$\times10^{-4}$ & 2.01$\times10^{-2}$ & 5.13$\times10^{-2}$ & 1.03$\times10^{-1}$ \\
\hline
70\%--30\% & 2.58$\times10^{-4}$ & 1.42$\times10^{-3}$ & 5.37$\times10^{-4}$ & 2.04$\times10^{-2}$ & 5.22$\times10^{-2}$ & 1.07$\times10^{-1}$ \\
\hline
\end{tabular}%
}
\caption{Average MSE and Relative L2 error for different train-test splits.}
\label{table:split_results}
\end{table}

As seen in Table \ref{table:split_results}, the performance metrics (MSE and Relative L2 error) for pressure, velocity, and turbulence remain very close across different training data fractions. This consistent performance across various training sizes demonstrates that the model is inherently robust, capable of achieving high performance even with varying amounts of training data. The model has effectively learned the underlying patterns of the dataset, ensuring reliable predictions. Each of the training datasets, from 70\% to 90\%, contains enough data for the model to generalize well, thus explaining the minimal performance differences.  The model's effective regularization techniques prevent overfitting, ensuring stable performance across different training sizes. These findings confirm that the model has been trained with a sufficient amount of data and performs well even with reduced training data. This robust performance implies that the model can be effectively used in scenarios with limited data availability. Moreover, the ability to maintain accuracy with varying training sizes indicates that the model is versatile and reliable. For the rest of the studies and demonstrations, we have chosen the 80-20 train-test split as it is a standard practice in the field, providing a good balance between training and testing data.\\

In a separate experiment, the train-test split was kept constant at 80\%-20\% while varying the number of nodes used for training. The original number of nodes was 11,340, and this was compared to a reduced number of 2,835 nodes which is one forth of the original number. The results, presented in Table \ref{table:nodes_results}, show that the model provided consistent performance even when the number of nodes was reduced by three-fourths.

\begin{table}[h]
\centering
\renewcommand{\arraystretch}{1}
\resizebox{\textwidth}{!}{%
\begin{tabular}{c|c|c|c|c|c|c}
\hline
\multirow{2}{*}{\textbf{Number of Nodes}} & \multicolumn{3}{c|}{\textbf{Average MSE}} & \multicolumn{3}{c}{\textbf{Average Relative L2 Error}} \\
\cline{2-7}
 & \textbf{Pressure} & \textbf{Velocity} & \textbf{Turbulence} & \textbf{Pressure} & \textbf{Velocity} & \textbf{Turbulence} \\
\hline
11340 & 2.55$\times10^{-4}$ & 1.40$\times10^{-3}$ & 5.28$\times10^{-4}$ & 2.01$\times10^{-2}$ & 5.13$\times10^{-2}$ & 1.03$\times10^{-1}$ \\
\hline
2835 & 2.68$\times10^{-4}$ & 1.42$\times10^{-3}$ & \textbf{5.07$\times10^{-4}$} & 2.07$\times10^{-2}$ & 5.14$\times10^{-2}$ & 1.03$\times10^{-1}$ \\
\hline
\end{tabular}%
}
\caption{Average MSE and Relative L2 error for different numbers of nodes.}
\label{table:nodes_results}
\end{table}

As shown in Table \ref{table:nodes_results}, the performance metrics (MSE and Relative L2 error) for pressure, velocity, and turbulence remain consistent across the different numbers of nodes. This suggests that the model's performance is not heavily dependent on the number of nodes used for training, indicating that the model is capable of maintaining accuracy even with fewer nodes. This robustness implies that the model can be effectively scaled down, which can be advantageous for computational efficiency without sacrificing predictive accuracy.

These two experiments demonstrate the consistency of the model's performance across varying training data sizes and different numbers of nodes, indicating its potential reliability in diverse scenarios. These findings highlight that the model is well-trained with the available data and can still perform effectively even with smaller datasets, making it practical for various applications and computationally efficient.\\

\subsection{Performance Evaluation} 

Building on our previous studies—comparing Vanilla DeepONet to DeepONet with Linear Layers, and analyzing the impact of data splits and node counts on model accuracy—we now focus on a comprehensive analysis of the modified DeepONet with linear layers using an 80-20 train-test split. This section provides a comprehensive evaluation of the model’s performance across the key parameters of turbulence kinetic energy, velocity, and pressure. Table \ref{table:model_metrics} summarizes the average Mean Absolute Error (MAE), Mean Squared Error (MSE), and Relative L2 error for the individual parameters on the test dataset. The results indicate that the model performs best in predicting pressure, as evidenced by the lowest values in MSE, MAE, and Relative L2 error. In contrast, while the MSE and MAE for velocity and turbulence remain relatively small, the model shows higher errors for velocity. Turbulence exhibits the worst performance in terms of Relative L2 error, with a relative error of 10.58\%.

\begin{table}[h!]
\caption{Average MAE, MSE, and Relative L2 error for individual parameters on the test dataset.}
\centering
\label{table:model_metrics}
\begin{tabular}{@{}c|c|c|c@{}}
\hline 
\textbf{Parameters} & \textbf{Average MSE} & \textbf{Average MAE} & \textbf{Average Relative L2 Error} \\ 
\hline 
Velocity      & \textbf{1.401$\times10^{-3}$}   & \textbf{3.1264$\times10^{-2}$} & 5.1348$\times10^{-2}$ \\ \hline 
Pressure      & 2.550$\times10^{-4}$            & 8.714$\times10^{-3}$           & 2.0108$\times10^{-2}$ \\ \hline 
Turbulence    & 5.280$\times10^{-4}$            & 1.6848$\times10^{-2}$           & \textbf{1.03441$\times10^{-1}$} \\ 
\hline 
\end{tabular}
\end{table}

To further illustrate the model’s performance, we provide a detailed error distribution analysis using histograms in Figure \ref{fig:mse_l2}. These histograms help us better understand the frequency distribution of MSE and Relative L2 error across the test dataset for each parameter, offering deeper insight into the variability and consistency of the model’s predictions. Panel (a) presents the distribution of MSE values for pressure predictions, indicating a concentration of lower error values, which suggests a strong model performance in pressure estimation. Similarly, panel (c) shows the model’s MSE distribution for velocity predictions, and panel (e) for the turbulence kinetic energy. The right-hand panels, (b), (d), and (f), illustrate the relative L2 error percentages. The embedded statistics within these panels (mean, standard deviation, and quantiles) summarize the error distribution, offering a comprehensive perspective on the model’s performance.

\begin{figure}[h!]
    \centering
    \includegraphics[width=0.95\textwidth]{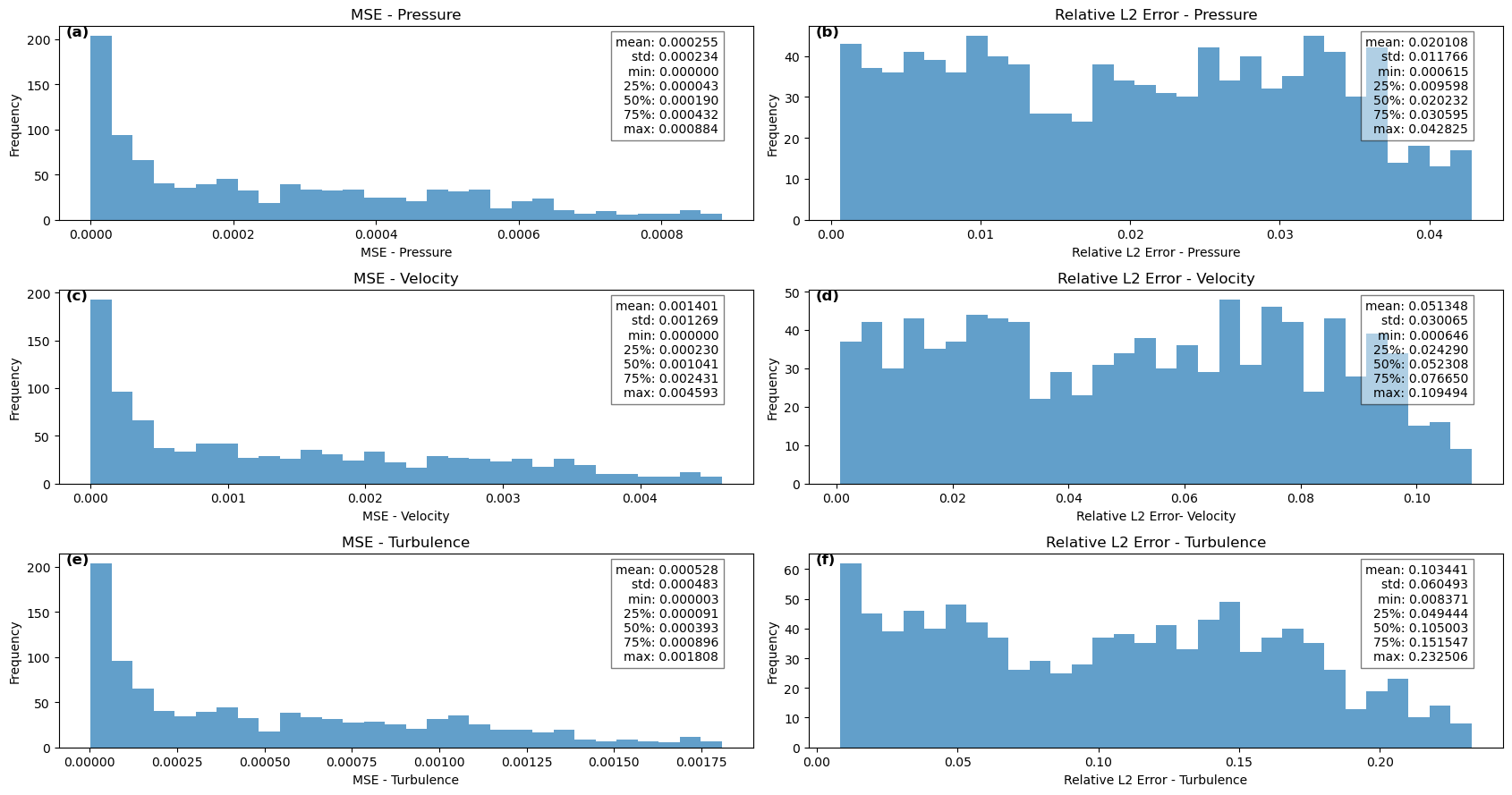}
\caption{Distribution of model performance metrics for DeepONet on the test dataset.}
\label{fig:mse_l2}
\end{figure}

The histograms for pressure predictions (Panels (a) and (b)) show a concentration of lower error values, indicating a strong and consistent model performance in estimating pressure. The narrow spread of MSE and Relative L2 error values reflects the model's ability to make accurate and reliable pressure predictions with minimal variability. For velocity predictions (Panels (c) and (d)), the MSE values exhibit a larger spread, suggesting greater variability in the model’s accuracy for this parameter. The higher MSE for velocity indicates that, on average, the errors in velocity predictions are larger compared to those in pressure predictions. However, the Relative L2 error for velocity is lower than that for turbulence, which means that, relative to the magnitude of the true values, the errors in velocity predictions are smaller compared to the errors in turbulence predictions. Turbulence kinetic energy predictions (Panels (e) and (f)) present the most significant challenge for the model. Although the MSE for turbulence is lower than that for velocity, the Relative L2 error is higher. This discrepancy signifies that while the mean square error in turbulence predictions might be smaller, the error relative to the actual values of turbulence is larger. In other words, the model's errors in predicting turbulence are more significant when viewed in the context of the scale of the turbulence values. This high Relative L2 error indicates that the model struggles more with predicting turbulence accurately compared to velocity and pressure, reflecting the complex nature of turbulence and the inherent difficulty in modeling it accurately.

A detailed visual analysis of the model's predictive performance, including best-case, worst-case, and transitional scenarios, is presented in Section \ref{vis} to further investigate the model's strengths and limitations.\\

\subsection{Inference Time}

DeepONet is well-known for its short inference time, especially when compared to traditional numerical methods such as the Finite Volume Method (FVM). After the model has been fully trained, a well trained DeepONet model can generate predictions almost instantaneously for new inputs. This is because it bypasses the need for iterative solving of differential equations, which is computationally expensive in traditional methods like FVM.

\begin{table}[h!]
    \centering
    \begin{tabular}{l|c|c}
     \hline
         & \textbf{FV Simulation Time [s]} & \textbf{Inference Time [s]}\\  
     \hline
     FV Simulation         & 200   & -    \\ 
     \hline
     DeepONet   &-     & 1.3507$\times10^{-1}$ \\  
     \hline
    \end{tabular}
    \caption{Comparison of inference time between DeepONet and FVM}
    \label{tab:inference_time}
\end{table}

Table \ref{tab:inference_time} shows the inference time for DeepONet in comparison to FVM simulation. As we can see, DeepONet is approximately 1481 times faster than FVM, which requires 200 seconds for a single simulation. This drastic reduction in prediction time makes DeepONet particularly useful for virtual sensing. This further reinforces our choice of utilizing DeepONet for this study.

\textbf{Note:} Both the FV simulation and DeepONet inference were performed on the same machine to ensure a fair comparison. The machine specifications were Intel Core i7 CPU with 16 GB RAM.

\subsection{Local Variable Analysis and Discussion} \label{vis}

To provide a deeper understanding of the model's predictive performance, we include a comprehensive visual analysis in Figure \ref{fig:vel_final}, \ref{fig:tke} and \ref{fig:pressure}. These figures showcase the ANSYS simulated data, the neural network's predictions, and the corresponding errors. For each parameter—pressure, velocity, and turbulence kinetic energy—we include five plots: the best-case scenario, the worst-case scenario, and three intermediate cases, all selected based on Relative $L2$ Error. The best-case plots highlight scenarios where the model's predictions closely align with FVM data, demonstrating optimal performance. Conversely, the worst-case plots reveal significant discrepancies and prediction errors. The intermediate plots show the gradual transition between these extremes, providing insights into regions where the model's performance deteriorates. This detailed visual analysis will not only help us understand which parameters the model predicts poorly but also identify the specific sections of the pipe where these inaccuracies occur. We can gain valuable insights into the model's limitations and highlight areas for refinement to enhance its overall accuracy and reliability.

Figure \ref{fig:vel_final}  presents the velocity contours for both the true and predicted values, along with the associated error distribution, on a selected diagnostic plane. The figure showcases the results for five representative cases, ranging from the best-performing case to the worst-performing case, with the intermediate cases illustrating a 25\% interval distribution of best performing case. The figure demonstrates that DeepONet, while generally proficient at predicting velocity patterns, encounters difficulties in regions with abrupt velocity transitions, particularly near the elbow joint where steep gradients and discontinuities are prevalent due to flow separation from the inner radius. In these areas, the model exhibits higher prediction errors compared to regions with smoother flow fields. This discrepancy can be attributed to the inherent challenge of capturing sharp gradients and discontinuities using a data-driven model like DeepONet. While the flow behavior in these zones is governed by intricate physical laws, DeepONet primarily relies on learning patterns from training data. Consequently, the model struggles to achieve high accuracy in regions where the flow physics plays a dominant role. This limitation is consistently observed across all five cases presented. \\

  \begin{figure}[htbp]
    \centering
    \includegraphics[width=1.0\textwidth]{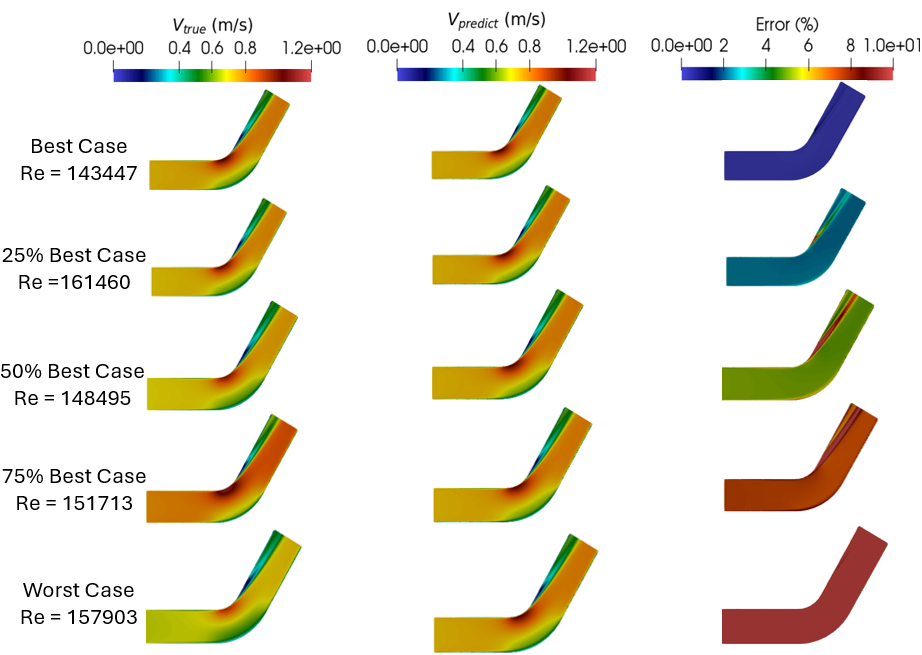}
    \caption{Visualization of true and predicted velocity at the focused plane.}
    \label{fig:vel_final}
\end{figure}  
  
Analysis of Turbulence Kinetic Energy (TKE) in Figure \ref{fig:tke} reveals that DeepONet accurately predicts turbulence patterns. However, error patterns for TKE are inversely related to those observed in velocity predictions. TKE magnitude increases downstream of the elbow due to flow separation and recirculation, resulting in pressure gradients and upsurged the turbulence. DeepONet excels at predicting local zones with high TKE values, but accuracy diminishes in regions of low TKE, even with smooth flow fields. This could be attributed to the nature of TKE magnitudes, which are near zero in smooth flows and significantly higher in turbulent regions. DeepONet struggles to capture these subtle variations near zero, leading to the observed error pattern reversal compared to velocity fields. Furthermore, the contour plots reveal instances of exceptionally high errors at specific nodes within the diagnostic plane, suggesting the presence of outliers in the predicted data. These outliers, characterized by significant deviations from the true values, could arise from various factors such as noise in the training data, model limitations in handling extreme or rare flow conditions, or numerical instabilities during the prediction process.\\

  \begin{figure}[htbp]
    \centering
    \includegraphics[width=1.0\textwidth]{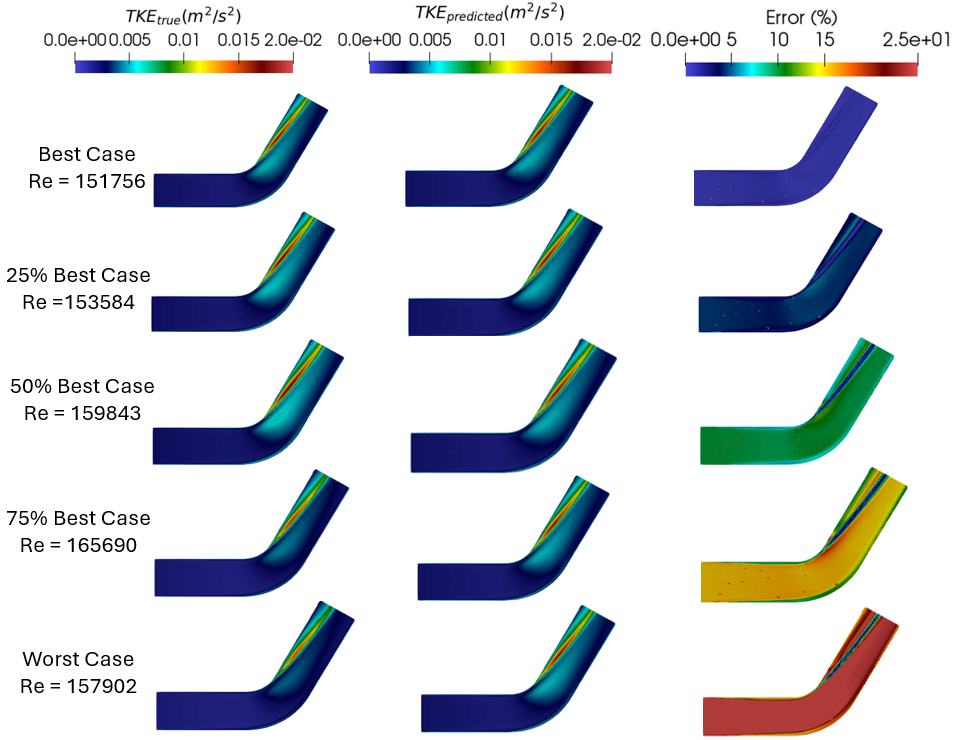}
    \caption{Visualization of true and predicted TKE at the focused plane.}
    \label{fig:tke}
\end{figure}  

Figure \ref{fig:pressure} represents the comparison of pressure between the CFD simulation and DeepOnet. In the $120^\circ$ elbow bend, centrifugal force causes the fluid to bulge towards the convex wall, decreasing flow rate and creating a high-pressure region. Conversely, the concave wall experiences contraction effects and increased velocity, resulting in a low-pressure zone. This pressure difference between the convex and concave sides drives a secondary flow from the convex to the concave wall along the bend's circumference.
The investigation into pressure field predictions echoes previous observations regarding DeepONet's performance in capturing velocity fields. While the model demonstrates general proficiency, it encounters challenges in regions characterized by abrupt pressure transitions, particularly near the elbow joint where steep gradients and discontinuities are prevalent. In these areas, the model exhibits elevated prediction errors compared to regions with smoother pressure distributions. This aligns with the inherent difficulty of capturing sharp gradients and discontinuities using a data-driven model like DeepONet. The pressure behavior in these zones is strongly influenced by complex flow physics, whereas DeepONet primarily learns patterns from training data.

  \begin{figure}[h!]
    \centering
    \includegraphics[width=1.0\textwidth]{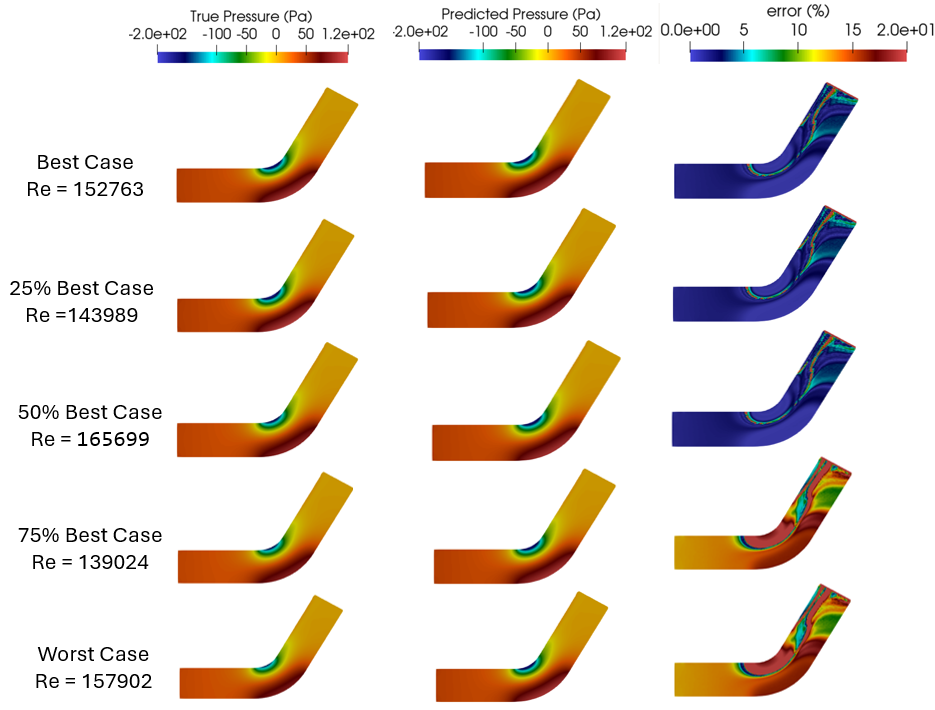}
    \caption{Visualization of true value and predicted Pressure in the focused plane.}
    \label{fig:pressure}
\end{figure}  

In general, the DeepONet model demonstrated a strong ability to accurately capture local fluctuations and patterns within the hydrodynamic variables. However, a disparity in prediction accuracy was observed across the different variables. Specifically, the highest errors were encountered in the turbulence kinetic energy predictions, followed by the pressure fields, with the velocity fields exhibiting the lowest errors. This suggests that the model's capacity to learn and generalize from the training data may be influenced by the inherent complexity and variability of the different flow variables. While DeepONet effectively captures the more readily observable velocity patterns, it struggles to achieve the same level of precision for the more nuanced and less directly measurable quantities such as turbulence kinetic energy and pressure. This could be attributed to the intricate physical relationships and nonlinear interactions governing these variables, which may not be fully captured by the current model architecture or training dataset.

Further research and model refinement, potentially incorporating additional physical constraints or specialized loss functions, may improve the accuracy of predictions for these more challenging flow variables.

\section{Discussion}

In this study, we have explored the integration of Digital Twin technology with advanced neural operator models, specifically DeepONet, to develop a virtual sensing-enabled framework for real-time parameter prediction in nuclear reactors. This section elaborates on the significance of our findings, the role of DeepONet within a Digital Twin framework, and the implications of our results for degradation monitoring. Additionally, we address the limitations of our approach and discuss potential future directions.

\subsection{DeepONet-based Digital Twin} 
The DeepONet developed in this study serves as a digital twin framework for plant operations, aligning with the U.S. Nuclear Regulatory Commission's (NRC's) definition of a digital twin through its integration of real-time inference, adaptability to operational condition changes, and synchronization with the physical system \cite{yadav2023state,liu2024development}.

\begin{itemize}
\item \textit{Dynamic Digital Representation and Synchronization with the Physical System:} DeepONet provides a dynamic representation of the system it models, offering real-time insights into the system’s behavior. Synchronization between the physical system and its digital representation is a key requirement for realizing digital twin technology. DeepONet addresses this need by processing real-time signals through a supporting data pipeline and dynamically updating its predictions, enabling seamless interaction with the physical system \cite{liu2024development}. DeepONet models can be trained using simulation data (such as ANSYS simulations used in this study) and later fine-tuned with real-world data, bridging the gap between digital and physical systems. This capability aligns with the key feature of digital twins to adapt and evolve alongside their physical counterparts. As shown in our study, the branch network processes real-time sensor data (average inlet velocity) ensuring synchronization, while the trunk network provides spatial details, functioning as a virtual sensor network. This setup allows DeepONet to augment physical sensor data by providing full-field predictions, such as flow velocity and turbulence, at critical locations like pipe bends and elbow joints, where physical sensors are impractical. The model's predictive capabilities, as shown in Figures \ref{fig:vel_final}, \ref{fig:tke}, and \ref{fig:pressure}, demonstrate its effectiveness in replicating system behavior. These results validate DeepONet’s role as a reliable and efficient component of the Digital Twin framework.

\item \textit{Real-Time Inference:} A key feature of digital twins is real-time or near-real-time inference to provide operational insights and enhance decision-making during dynamic conditions. As shown in Table \ref{tab:inference_time}, DeepONet achieves unprecedented speed, making predictions 1481 times faster than traditional finite volume simulations. This allows operators to monitor critical thermal-hydraulic parameters—pressure, velocity, and turbulence kinetic energy—almost instantaneously. The real-time inference capability ensures the model remains an active, dynamic component of the digital twin framework, crucial for nuclear plant operations where quick decision-making is essential.

\item \textit{Adaptability to Operational Condition Changes:} Digital twins must adapt to changing operational conditions and maintain accuracy across diverse scenarios without constant retraining. This study demonstrates that DeepONet fulfills this requirement by learning the underlying functional relationships between input parameters and spatial coordinates. Its architecture ensures robust predictions under varying conditions, without requiring retraining. As shown in Figure \ref{fig:mse_l2}, DeepONet performs consistently across diverse scenarios. It effectively predicts spatial distributions of velocity, turbulence, and pressure across different inlet conditions, showcasing its adaptability and reliability for monitoring dynamic systems. This capability aligns with the NRC’s definition of digital twins, emphasizing adaptability and synchronization with evolving physical systems.
\end{itemize}

These findings validate DeepONet's role in enabling a Digital Twin framework, bridging real-time data and predictive modeling.

\subsection{Monitoring Operational Conditions Indicative of Degradation}

Degradation, such as material/wall thinning, stress corrosion cracking, or fatigue, develops gradually due to persistent mechanical stresses and flow irregularities. Certain operational conditions, however, may act as early indicators of these processes and can be monitored to prevent long-term damage. For instance, flow-induced vibrations, caused by turbulent flow around pipe bends and fittings, are a major contributor to fatigue and erosion-corrosion in nuclear piping systems. These vibrations can lead to high-frequency stresses, overloading pipe supports or nozzles and accelerating material degradation. Similarly, transient phenomena such as water hammer or condensation-induced hammer, generate impact loads with high dynamic factors, which can weaken pipe walls or connections over time. Monitoring turbulence pattern, pressure fluctuations, and velocity distributions in such areas provides actionable insights into regions where degradation risks are elevated. \cite{Jacimovic2020}

While our study does not directly simulate material degradation or wall thinning, the ability to monitor turbulence and pressure in areas prone to stress highlights DeepONet's usefulness in \textit{degradation-informed decision-making}. Elevated turbulence near pipe bends is a known precursor to erosion-corrosion, while fluctuating pressure often signals transient phenomena such as water hammer, both of which can severely impact system integrity. So say for example, a rise in turbulence intensity near elbow joints (Figure \ref{fig:tke}) over time could signal heightened mechanical stress in those regions, suggesting targeted inspections or preventive maintenance. Similarly, deviations in pressure (as in Figure \ref{fig:pressure}) or velocity distributions (shown in Figure \ref{fig:vel_final}) can identify developing flow-induced vibrations that may accelerate wear, particularly in locations vulnerable to acoustic resonance. 

Within a Digital Twin, these data can be collected over time to form flow profiles, revealing how conditions within a pipe evolve dynamically. Such profiles are invaluable for identifying patterns indicative of emerging degradation risks, such as increasing pressure in certain region or abnormal turbulence patterns near pipe bends. Furthermore, by illuminating unmonitored regions with accurate predictions, DeepONet enhances the visibility of the entire system, enabling reliable condition-based maintenance strategies, that can reduce maintenance cost and enhance reactor safety.

\subsection{Limitations and Future Directions}

While the results of this study show the effectiveness of DeepONet in predicting key thermal-hydraulic parameters, certain limitations must be acknowledged. One notable challenge is the spectral bias inherent in data-driven models like DeepONet. All Neural networks tend to prioritize learning low-frequency, smoother patterns over high-frequency, more complex ones, such as those observed in turbulent regions. This spectral bias, coupled with the imbalanced nature of our dataset, evident in Figures \ref{fig:vel_final}, \ref{fig:tke}, and \ref{fig:pressure}, where larger, smoother flow patterns dominate over the relatively smaller turbulent flow patterns in the bend region, likely contributed to the increased error observed in predicting turbulence. We assume that the model is smoothing out high-frequency patterns. For example, the model exhibited higher errors in areas with intense turbulence, such as near elbow joints and bends, where high-frequency flow patterns dominate. These regions are most important for identifying flow-induced vibrations and other precursors to degradation, highlighting a potential limitation of the current approach.

To address this, future research could explore combining neral operator with diffusion models \cite{Oommen2024}. Future directions could also include designing hybrid frameworks, where specialized models can be trained for different regions of the system based on the initial CFD modeling. This targeted approach could improve the performance by allowing each model to focus on specific flow characteristics, thereby addressing challenges such as spectral bias and ensuring better accuracy in regions with high-frequency turbulent flows.

\section*{Acknowledgment}
The authors were supported by the National Science Foundation (award OAC 2005572 for Delta supercomputing facility), the United States Department of Energy (DOE), and Battelle Energy Alliance LLC (Idaho National Lab).

\section*{Data and code availability}
The data and code used and/or analyzed during this study are available from the corresponding author on reasonable request.

\section*{Competing Interests}
The authors declare no conflict of interest.

\section*{Author Contributions}
Conceptualization: S.A., R.H., F.A. and K.K.; methodology: R.H. and F.A.; Writing - original draft: R.H. and F.A.; Review \& Editing: All authors; Formal analysis: R.H.; Visualization: R.H. and F.A.; Data curation: R.H. and F.A.; Supervision: S.A.; Project administration: S.A. and D.A.; All authors have read and agreed to the published version of the manuscript.

\section*{Declaration of Generative AI and AI-assisted Technologies in the Writing Process}
During the preparation of this work, the author(s) utilized a Large Language Model (ChatGPT) to assist with language editing and refinement. The AI tool was employed strictly to enhance the clarity, grammar, and overall readability of the manuscript, ensuring the effective communication of ideas. It is important to note that no scientific content, original ideas, or conceptual contributions were generated, altered, or influenced by the AI.

\bibliographystyle{unsrtnat}
\bibliography{references} 

\end{document}